\documentclass[10pt]{article}
\usepackage{geometry} % see geometry.pdf on how to lay out the page. There's lots.
\pdfoutput=1
\usepackage{caption}
\usepackage{amssymb}
\usepackage{amsthm}
\usepackage{amsmath}
\usepackage{mathtools}
\usepackage{bbm}
\usepackage{mathenv}
\usepackage{hyperref}
\usepackage{wrapfig}
\usepackage{subcaption}
\usepackage{graphicx}
\usepackage{enumitem}
\usepackage[utf8]{inputenc}
\usepackage[english]{babel}
\usepackage{mathtools}
\usepackage{verbatim}
\usepackage{color}
\usepackage{todonotes}

\begin{document}

\begin{center}
\large{\textbf{Inferring the time-varying functional connectivity of large-scale computer networks from emitted events\footnote[2]{This work is supported by Moogsoft Ltd. and describes patented features of its products.} }}\\
\vspace{0.5cm}

\textit{Antoine Messager$^{1}$, George Parisis$^{1}$, Istvan Z. Kiss$^{2}$, Robert Harper$^{3}$, Phil Tee$^{4}$ and Luc Berthouze$^{1,*}$}
\end{center}
\vspace{0.0cm}
\begin{center}
$^1$ Department of Informatics, University of Sussex,
Falmer BN1 9QH, UK\\
$^2$ Department of
Mathematics, University of Sussex, Falmer BN1 9QH, UK\\
$^3$ Moogsoft Ltd, 31-35 High St, Kingston upon Thames KT1 1LF, UK\\
$^4$ Moogsoft Inc, 1265 Battery St, San Francisco, CA 94111, USA
\end{center}
%\vspace{1.5cm}
\begin{center}
\textbf{Abstract}\\
\end{center}
We consider the problem of inferring the functional connectivity of a large-scale computer network from sparse time series of events emitted by its nodes. We do so under the following three domain-specific constraints: (a) non-stationarity of the functional connectivity due to unknown temporal changes in the network, (b) sparsity of the time-series of events that limits the effectiveness of classical correlation-based analysis, and (c) lack of an explicit model describing how events propagate through the network. Under the assumption that the probability of two nodes being functionally connected correlates with the mean delay between their respective events, we develop an inference method whose output is an undirected weighted network where the weight of an edge between two nodes denotes the probability of these nodes being functionally connected. Using a combination of windowing and convolution to calculate at each time window a score quantifying the likelihood of a pair of nodes emitting events in quick succession, we develop a model of time-varying connectivity whose parameters are determined by maximising the model's predictive power from one time window to the next. To assess the effectiveness of our inference method, we construct synthetic data for which ground truth is available and use these data to benchmark our approach against three state-of-the-art inference methods. We conclude by discussing its application to data from a real-world large-scale computer network. \newline

%\vspace{1cm}

\noindent $^{*}$ Corresponding author: l.berthouze@sussex.ac.uk

\section{Introduction}

To better understand real world complex systems, it is often useful to describe them in terms of their mesoscopic or macroscopic features. Many systems including the Internet, gene regulatory systems and brains can be described as networks where nodes represent components and edges represent relationships between those components. Such relationships can take various forms. They could denote the existence of a structural link between the nodes (for example, synaptic connectivity between neurons) or that of a functional link (for example, two distant brain areas synchronising during the performance of a task). Although the term functional connectivity is being primarily used in neuroscience~\cite{sporns2010networks}, it can easily be applied to various other domains. In computer networks, structural connectivity would describe the presence of physical links~\cite{calvert1997modeling} between nodes (e.g., servers, routers, switches), whereas functional connectivity may describe the integrated involvement of these different nodes in the provision of a particular service. Services may be realised at different network layers by a potentially large number of in-network and edge devices, e.g., a set of routers that form an OSPF area, a set of switches that are part of a spanning tree, an application deployment that consists of application and database servers, load balancers and a firewall. This paper is concerned with inferring the functional connectivity of a computer network from events emitted by its nodes, where functional connectivity refers to the set of connectivities corresponding to distinct services. 

Although structural connectivity will, in most cases, underpin functional connectivity, it is not always possible to identify structural links. In brain networks, even the most advanced forms of imaging cannot provide an accurate or complete description of structural connectivity, see~\cite{thomas2014anatomical} for example. In the case of Internet, its size, complexity and anarchistic evolution combined with its distribution over many private Internet Service Providers (ISPs) render topological mapping very difficult~\cite{bajpai2015survey}. Although link prediction methods are being developed to attempt to extract missing information~\cite{liu2016network,lu2011link}, a more promising approach is to infer connectivity from the temporal evolution of events occurring at node level~\cite{brugere2016network} without assuming any prior knowledge regarding connectivity. In brain neural networks, for example, network inference may rely on spike dynamics~\cite{brown2004multiple}. In gene regulation, published methods primarily use gene expression data derived from micro-arrays~\cite{hecker2009gene}.  

In this paper, we consider the problem of inferring the functional connectivity of a large-scale computer network from time series of node events. This particular application domain (described in Section~\ref{subsec:rwds}) imposes three important constraints on the development of a suitable inference method: (a) non-stationarity of the functional connectivity due to unknown temporal changes in the network, (b) sparsity of the time-series of events that can limit the effectiveness of classical correlation-based analysis, and (c) lack of an explicit model describing how events propagate through the network.  
Since we are not aware of any method that simultaneously deals with all three constraints, in what follows, we briefly describe those works that handle some combination of them. 

A number of methods have been developed to infer connectivity changes in network by adapting Bayesian network methods, see~\cite{dondelinger2013non,wang2011time,robinson2010learning,song2009time} for examples. These methods assume a model of event  propagation. At their core is the belief that the current state can be predicted, with some probability, based on the previous state or states. These methods assume that if there is an edge between A and B, then if A emits an event, B is very likely to also emit an event. If this succession of events does not occur, it is attributed to noise. In the context of very sparse data when an event in A may trigger an event in B in only a small percentage of the cases, such an assumption is problematic, especially if the signal noise ratio cannot be determined. A consequence of the sparsity constraint is that one cannot assume that events propagate to all neighbouring nodes throughout the recording. For this reason, methods such as the Markovian model used in temporal exponential graphs~\cite{guo2007recovering}, various adaptations of the Kalman filter~\cite{carluccio2017akron,khan2014tracking} or methods relying  on propagation of cascades~\cite{rodriguez2014uncovering} are unlikely to be as effective as a method that will primarily rely on pair-wise information.

The use and adaptation of pair-wise correlations is the basis of many methods that do not assume an event propagation model, see ~\cite{Oliner2010,Mahimkar2009,Zheng2012} for examples in the computer network domain. However, these adaptations typically result in methods that do not scale well to large networks over long recordings. In Section~\ref{subsec:benchmark}, we will introduce our own adaptation and use it as one of 3 benchmark methods. Another approach to measuring pair-wise interactions is that of Kobayashi et al.~\cite{Kobayashi2017}. Their method is well adapted to sparse data and, although it is not specifically aimed to infer changing topology, it can be used to do so. That is what we do in Section~\ref{subsec:benchmark} and the method will be described in more detail then. 

A final class of methods rely on the estimation of a time-varying covariance matrix to encode the correlation structures at each observation, e.g.,~\cite{monti2014estimating,wit2015inferring,zhou2010time}. Constraints of sparsity (in the network of interdependencies between the nodes) are enforced by way of lasso penalty. These methods typically do not scale well to large examples. Hallac et. al~\cite{hallac2017network} recently proposed a more scalable implementation. We attempted to use this method as final benchmark in Section~\ref{subsec:benchmark}, where we describe it in more detail.  

The paper is organised as follows. In Section~\ref{sec::methods}, we describe the proposed methodology. In Section~\ref{sec::results}, we validate the method by applying it to synthetic data for which ground truth regarding functional connectivity is available. We then demonstrate its effectiveness in dealing with our three constraints by benchmarking it against the three methods identified above. After applying it to data from a real-world large scale computer network, we conclude by discussing limitations and possible avenues for further work.

\section{Methods}\label{sec::methods}
Large-scale computer networks typically generate a very large number of events, with event rates at network level up to $10^6$ events per second. However, depending on how systems are configured, event rates at node level can be extremely low. For example, the real-world case scenario that will be discussed in Section~\ref{subsec:rwds} involves a mean waiting time between events (per node) of $10^5s$ on average (with a sampling time of $1s$). 
This extreme sparsity makes any estimation of the correlation structure difficult due to vanishing means. Therefore, the fundamental empirical assumption underlying this work is that the probability of two nodes being functionally connected correlates with the mean delay between their respective events. In the following subsection, we introduce a windowed measure of the temporal relationship between the events emitted by two nodes. This measure (referred to as score thereafter) will then be used to build a model of time-varying edge probabilities (which will be described in Section~\ref{sec::model}). 

%%%%%%%%%%%%%%%%%%%%%%%%%%%%%%%%%%%%%%%%%%%%%%%%%%%%%%%%%%%%%%%%%%%%%%%%%%%%%%%%%%%%%%%%%%%%%%%%%%%%%%%%%%%%%%%%%%%%
\subsection{Score: estimating pairwise functional couplings}\label{sec::score}
%%%%%%%%%%%%%%%%%%%%%%%%%%%%%%%%%%%%%%%%%%%%%%%%%%%%%%%%%%%%%%%%%%%%%%%%%%%%%%%%%%%%%%%%%%%%%%%%%%%%%%%%%%%%%%%%%%%%

We consider sampled time series where the value at each observation denotes the presence of a node event (at most one per observation by construction). To quantify the presence of recurring temporal interaction between two nodes in the interval $[0;T]$, we calculate the cross-correlation of their respective time series $f$ and $g$. 
Specifically, as we do not assume a spreading process, we use an adaptation of the cross-correlation that does not distinguish between positive and negative delays: 
\begin{equation}
(f\star g)(\delta t) = \sum_{t=\delta t}^{T - \delta t} f(t) \Big(g(t +  \delta t) + g(t -  \delta t)\Big),
\end{equation}
where $\delta t$ denotes a lag in units of sample time. 

In a large-scale network, calculating cross-correlations over all possible pairwise interactions is computationally extremely intensive and therefore in our implementation cross-correlations were only calculated up to a maximum lag of $\tau_{max}$. This value was set based on an analysis of the histogram of co-occurrences for a given lag and was such that for any larger lag, the number of co-occurrences was negligible.

To determine the presence of a functional link between two nodes, we construct a score that characterises the shape of the cross-correlation in terms of the number of its peaks and the relative distribution of these peaks within a range of delays. Namely, it is assumed that two nodes are likely to be functionally connected if there are multiple peaks in their cross-correlation (assumption 1) and there are more peaks for small delays than for large ones (assumption 2).
To translate the first assumption into a quantitative measure, we calculate $R_{f,g}(\tau)$, the number of peaks in the cross-correlation for all lags in the interval $0,\tau$ where $\tau$ is a delay in units of sample time. It is given by: 
\begin{equation}
\label{eq:R}
R_{f,g}(\tau) = \sum_{\delta t=0}^{\tau} (f\star g)(\delta t). 
\end{equation} 

It is important to note that this quantity is sensitive to the number of events emitted by $f$ and $g$ and therefore does not support valid comparisons between different pairs. In what follows, we consider that two pairs of nodes with event time series $(f_1,g_1)$ and $(f_2,g_2)$ belong to the same grouping if the products of their number of events are approximately equal: $n_{f_1}\times n_{g_1} \approx n_{f_2}\times n_{g_2}$\footnote{Intuition for this condition comes from noting that for two independent Poisson processes, the expected cumulated number of peaks is proportional to the product of their number of events}. In our implementation, this quantity is binned in order to obtain a computationally tractable number of groupings with each grouping featuring a sufficiently high number of pairs. 

To quantify assumption 2 (the propensity of functionally linked nodes to have more peaks at smaller delays), we define $s(f,g)$ (referred to as score henceforth) as the maximum over all delays $\tau$ (with $\tau \in [0,\tau_{max}]$) of the deviation of the cumulative number of peaks $R_{f,g}(\tau)$ %in the cross-correlation between $f$ and $g$ 
from the mean number of peaks $R_{i,j}(\tau)$ for all possible pair of nodes in the grouping normalised by the standard deviation of the $R_{i,j}(\tau)$. Namely: 
\begin{equation} 
	s(f,g) = \max_{\tau}\Big(\frac{R_{f,g}(\tau) - \mu(\tau) }{\sigma(\tau)}\Big),
\end{equation}
where $\mu(\tau)$ and $\sigma(\tau)$ denote the mean and standard deviation of $R_{i,j}(\tau)$ for all possible pair of nodes $(i,j)$ in the grouping. 

As the cumulative number of peaks $R_{f,g}(\tau)$ is an increasing function of the delay $\tau$, both mean $\mu(\tau)$ and standard deviation $\sigma(\tau)$ are also increasing functions of $\tau$. The smaller the delay between two events is, the greater the ratio will be. Note that since we cannot make any assumption regarding the distributions of the cumulated number of peaks at a given delay, we refer to the ratio as normalised deviation rather than Z-score. 
 
\begin{figure}[h]
\begin{center}
    \includegraphics[width=0.75\linewidth]{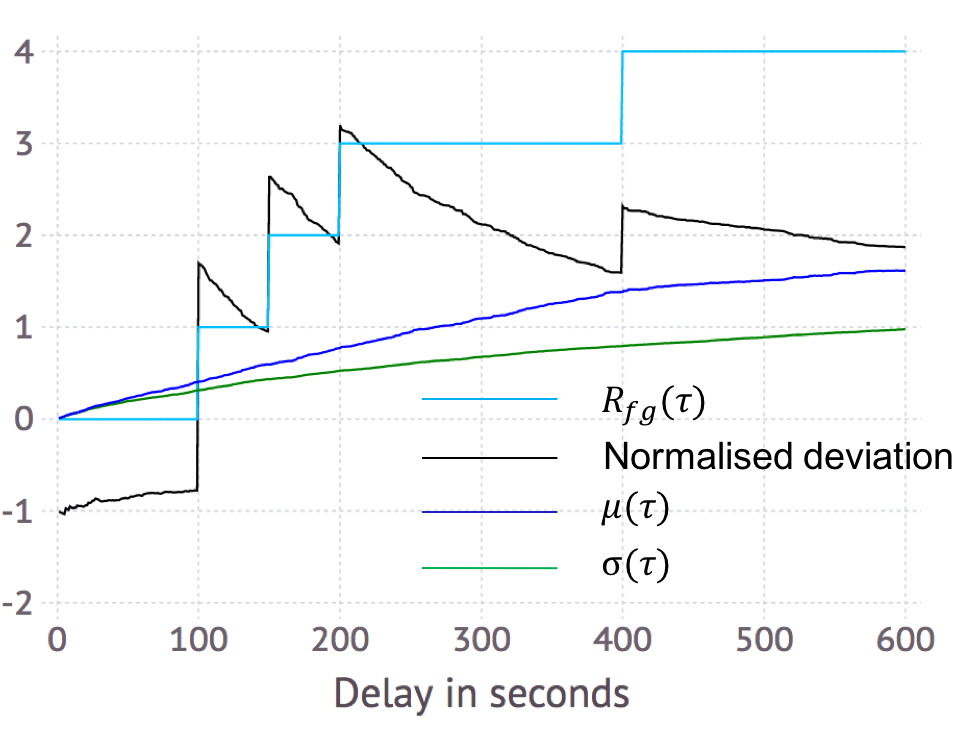}
    \caption{Determination of the score for a pair of nodes with event times series $f$ and $g$. The cross-correlation has peaks at time 100, 150, 200 and 400 leading to the cumulated number of pairs shown in cyan. The mean $\mu$ and standard deviation $\sigma$ of this quantity over all pairings in the grouping to which the nodes belong is shown in blue and green respectively. This enables the calculation of the normalised deviation (in black). The score $s(f,g)$ is defined as the maximum of this normalised deviation, here, $3.1$ at $\tau=200$.} 
    \label{fig:score_example}
    \end{center}
\end{figure}

Figure~\ref{fig:score_example} illustrates the process of determining the score for a pair of nodes. 

%%%%%%%%%%%%%%%%%%%%%%%%%%%%%%%%%%%%%%%%%%%%%%%%%%%%%%%%%%%%%%%%%%%%%%%%%%%%%%%%%%%%%%%%%%%%%%%%%%%%%%%%%%%%%%%%%%%%
\subsection{Model of time-varying connectivity}\label{sec::model}
%%%%%%%%%%%%%%%%%%%%%%%%%%%%%%%%%%%%%%%%%%%%%%%%%%%%%%%%%%%%%%%%%%%%%%%%%%%%%%%%%%%%%%%%%%%%%%%%%%%%%%%%%%%%%%%%%%%%

In this section, we describe our approach to translating the scores introduced in Section~\ref{sec::score} into time-varying probabilities of the existence of functional edges. 
We assume that for a suitably chosen time window, changes in connectivity will be small enough that the cross-correlations (and therefore the scores) can be considered stationary within that time window. In the context of network management, this assumption is realistic. In a large-scale computer network, the structural connectivity changes as result of failing devices (e.g., servers, routers) or when hardware is commissioned / de-comissioned. In both cases, the functional topology can be expected to also change. Failures do happen frequently but result in a stream of network events emitted by neighbouring or monitoring devices (at the level of structural or functional connectivity) that is exploited by the proposed inference method. Commissioning / de-commissioning of hardware or network services is usually planned and expected to take place within a short and known time frame. In Section~\ref{sec::results}, a range of time windows will be considered. 
\begin{figure}[h]
    \includegraphics[width=\linewidth]{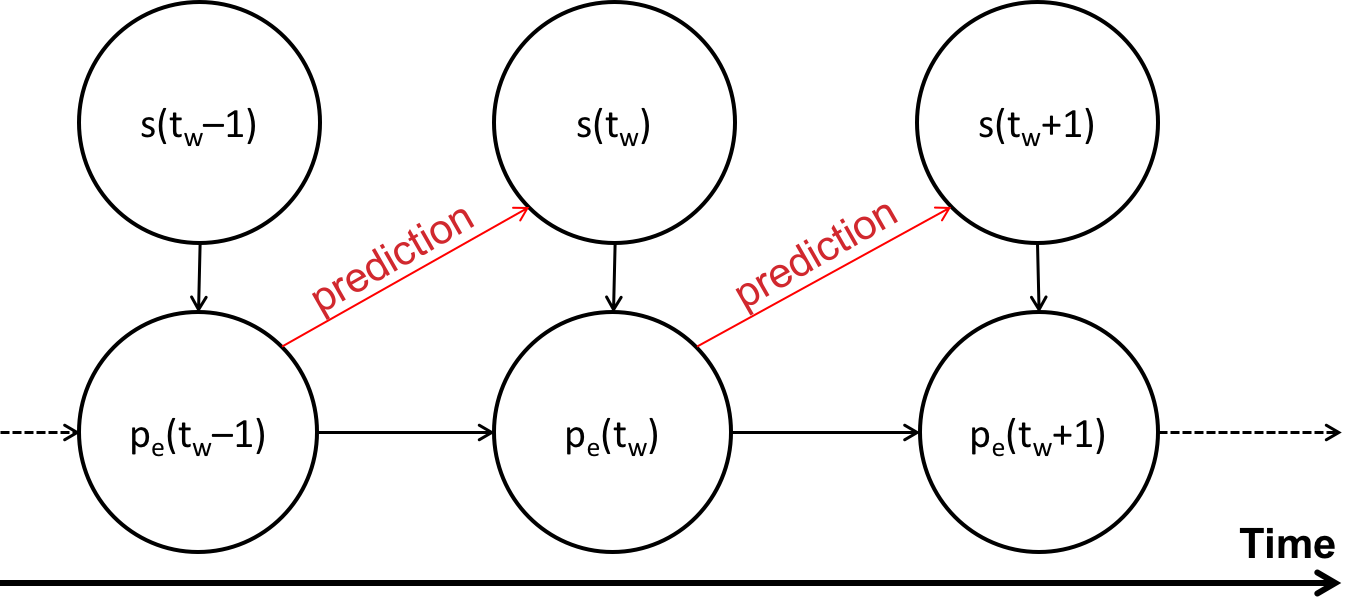}
    \caption{Iterative involvement of scores and edge probabilities in the proposed model. See details in text.} 
    \label{fig::principle}
\end{figure}

A key principle of the proposed methodology is that the score $s(t_w)$ for a pair of nodes within a time window $t_w$ provides the information required to update the estimate of the value of the probability $p_e(t_w-1)$ of a functional edge existing between these nodes at the previous time window (see Figure~\ref{fig::principle}). More precisely, we consider that information is gained about the probability of an edge existing only when both nodes emit events during the time window considered. This is a natural implication of the sparsity constraint. The fact that only one node in a pair emits an event does not necessarily imply that an edge does not exist (or no longer exists). For each pair of nodes and each time window $t_w$, there are therefore three cases to consider: 

\begin{enumerate}
\item The score is higher than expectation, $s(t_w) > 0$: This increases confidence about the existence of an edge and therefore the probability $p_e(t_w)$ should increase as some function $h$ of the score. 
\item The score is lower than expectation, $s(t_w) \leqslant 0$: This lowers confidence about the existence of an edge and therefore the probability $p_e(t_w)$ should decrease. 
\item At least one of the node does not emit events: This scenario does not provide any information and the probability should remain unchanged.
\end{enumerate}

Finally, we consider that as time passes, previous information loses currency and therefore a decay should be introduced. 
This leads to the following model formulation: 
\begin{equation}
p_e(t_w+1) = d \times 
 \begin{cases}
   \begin{aligned}
       & \Big( 1 - (1 - p_e(t_w))   \\ %{Dorogovtsev:2008gv}
       &\times(1-h(s(t_w))) \Big)
    \end{aligned}   
       & \text{\textbf{if} $s(t_w) > 0$}, \\
    & \\
    k \times p_e(t_w)  
     & \text{\textbf{if} $s(t_w)  \leqslant 0$},   \\
    & \\
    p_e(t_w) &  \text{\textbf{if} no information},
\end{cases}
\label{eq:dynamics}
\end{equation}
where $d$ and $k$ are decay parameters in $[0;1]$. If $d=1$, all past information is retained. Our implementation of case 1 (i.e., when the score is high) ensures that $p_e(t_w)$ remain bounded by 0 and 1. It involves decreasing the probability of not having an edge ($1-p_e(t_w)$) by a factor $(1-h(s(t_w)))$ where $h : \ ]0;\infty[ \rightarrow [0;1]$ is a continuous monotonic function of the score. 

In our implementation, $h$ was defined as: 
\begin{equation}
h(s(t_w)) = 
\begin{cases}
   0& \text{\textbf{if} $\alpha + \beta \ log(1 + s(t_w)) < 0$},\\
    \alpha + \beta \log(1 + s(t_w))\eqqcolon v &\text{\textbf{if} $ 0 \leq v \leqslant 1$}, \\  
     1& \text{\textbf{if} $\alpha  + \beta \ log(1 + s(t_w)) > 1$},
\end{cases}
\end{equation}
with $\alpha$ and $\beta$ two positive real numbers. The choice of $h$ as an affine function of the exponent of the score was heuristic and followed from the observation that since scores display a very wide range of positive values (from 0.5 to $10^4$), the exponent of the score would make for a more meaningful quantity. Other formulations are possible, and provided they are differentiable in their parameters (here, $\alpha$ and $\beta$), do not affect the principle of the method. It will be noted that the probability update equation when scores are negative does not involve any function of the score but is modelled as a simple decay. This is due to the absence of any empirical intuition. Altogether, the model involves four parameters ($\alpha$, $\beta$, $d$ and $k$) that need to be determined. 

Since changes in functional connectivity from one window to the other are assumed to be small, we formulate the problem of determining these four parameters as one of minimising the error of a binary classifier predicting the sign of the score at time $t_w$ given the edge probability at time $t_w-1$. In other words, if the edge probability at time $t_w-1$ is greater than a threshold $th$ and both nodes emit events in time window $t_w$, we expect the score at time $t_w$ to be positive. Conversely, if the edge probability at time $t_w-1$ is less than the threshold and both nodes emit events in time window $t_w$, we expect the score at time $t_w$ to be negative. Our proposed error criterion is formally defined as follows: 
\begin{equation}
\begin{split}
E = \sum_{t_w=1}^{N_w}\Bigg(\quad \quad \quad \quad & \sum_{\mathclap{s(t_w) \leqslant 0  \text{  and  } p_e(t_w-1) \geqslant th}} (p_e(t_w-1) - th) \quad + \\
&\sum_{\mathclap{s(t_w) > 0  \text{  and  } p_e(t_w-1) <  th}} (th - p_e(t_w-1))\Bigg). \\
\end{split}
 \label{eq:error}
 \end{equation} 

It penalises misclassifications, namely $p_e(t_w) \geqslant th$ and $s(t_w) \leqslant 0$, or $p_e(t_w) > th$ and $s(t_w < 0)$), with a cost proportional to the difference between edge probability and threshold. As a sum of edge probabilities that are differentiable functions of the parameters, a simple gradient descent can be used to determine the values of parameters $\alpha$, $\beta$, $d$ and $k$. Time-varying edge probabilities can then be calculated for all pairs using the update equation~(\ref{eq:dynamics}).

\section{Experimental results}\label{sec::results}

\noindent Validating a method that infers \textit{functional} connectivity presents a number of challenges because more often than not no ground truth is available. In what follows, we begin by demonstrating the effectiveness of our proposed methodology in handling sparse time series and changing functional connectivity by testing it against synthetic data (Section~\ref{subsec:synthdata}) and comparing it against the state-of-the-art approaches of Hallac et al.~\cite{hallac2017network} and Kobayashi et al.~\cite{Kobayashi2017}, as well as an adaptation of the classical correlation-based approach (Section~\ref{subsec:comparison}). We then discuss its application (Section~\ref{sec::RWapp}) to the data from our real-world application (Section~\ref{subsec:rwds}). 

%%%%%%%%%%%%%%%%%%%%%%%%%%%%%%%%%%%%%%%%%%%%%%%%%%%%%%%%%%%%%%%%%%%%%%%%%%%%%%%%%%%%%%%%%%%%%%%%%%%%%%%%%%%%%%%%%%%%
\subsection{Validation and benchmarking}
%%%%%%%%%%%%%%%%%%%%%%%%%%%%%%%%%%%%%%%%%%%%%%%%%%%%%%%%%%%%%%%%%%%%%%%%%%%%%%%%%%%%%%%%%%%%%%%%%%%%%%%%%%%%%%%%%%%%
 
%%%%%%%%%%%%%%%%%%%%%%%%%%%%%%%%%%%%%%%%%%%%%%%%%%%%%%%%%%%%%%%%%%%%%%%%%%%%%%%%%%%%%%%%%%%%%%%%%%%%%%%%%%%%%%%%%%%%
\subsubsection{Construction of the synthetic data}\label{subsec:synthdata}\

\noindent We generated synthetic data such that ground truth on underlying functional connectivity was available and further that the parameters of this functional connectivity could be manipulated and enable sensitivity analysis. 

\noindent The functional connectivity was defined as a partition\footnote{We make no assumption regarding connectivity within each functional group.} of a set of $N$ nodes into $n_{fg}$ functional groups, with (initially) $n_{dev}(fg)$ devices per functional group. 
Over the duration of the recording $[0,T]$, each functional group experienced a number $n_{casc}(fg)$ of cascades of events. The times $t_{casc}(fg)$ at which the cascades started were chosen uniformly at random (which corresponds to a homogeneous Poisson process). Each cascade affected a proportion $per_{dev}$ of devices in the functional group. These affected devices emitted events at times $t_{casc}(fg)+d_{dev}$ where $d_{dev}$ was chosen uniformly at random in the interval $[0,d_{max}]$ with $d_{max}$, the maximal delay possible. Because we observed that within the real-world dataset (see Section~\ref{subsec:rwds}), the number of events emitted by each device followed a scale-free like distribution, a preferential attachment mechanism was used to allocate events to nodes. Therefore instead of uniformly selecting a proportion of devices at each cascade, we selected devices with a probability proportional to the number of events it had already emitted. Finally, to simulate changes in the network, we set a number $n_{step}$ of times when $n_{change}$ devices changed functional group (which led to fluctuating numbers $n_{dev}(fg)$ of devices per functional group). The steps were evenly distributed within the time recording. 

%%%%%%%%%%%%%%%%%%%%%%%%%%%%%%%%%%%%%%%%%%%%%%%%%%%%%%%%%%%%%%%%%%%%%%%%%%%%%%%%%%%%%%%%%%%%%%%%%%%%%%%%%%%%%%%%%%%%
\subsubsection{Measure of accuracy}\

\begin{figure}[h]
\begin{center}
    \includegraphics[width=\linewidth]{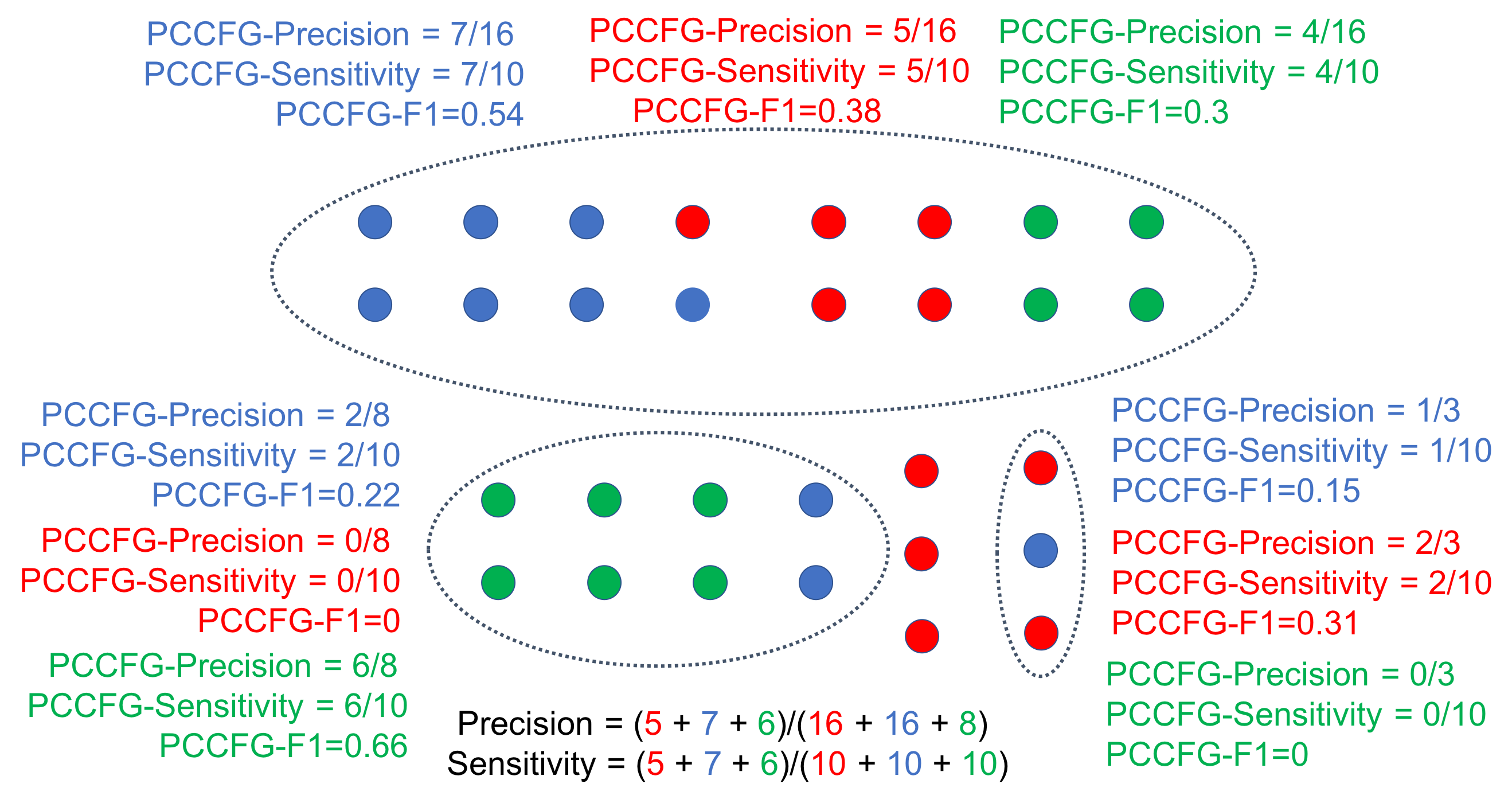}
    \caption{Illustration of the accuracy measure. The ground truth functional connectivity consists of 3 functional groups of 10 nodes each (green, red and blue). The reader should recall we do not assume any connectivity between those nodes. In this instance, the method considered inferred a functional connectivity consisting of 3 connected components, identified by the groupings shown in black dashed lines. For each of them, three PCCFG-F1 scores are computed for each of the 3 functional groups (coloured in blue, red and green). This results in matching the largest component with both the blue and red functional groups, and the second largest component with the green functional group. The overall precision and accuracy is calculated on the basis of 7 blue nodes, 5 red nodes and 6 green nodes being identified to connected components of size 16, 16 and 8 nodes respectively.} 
   \label{fig:accuracy}
   \end{center}
\end{figure}

\noindent Since we imposed no connectivity within the functional group, we assessed the accuracy of the inferred networks on a node (rather than edge) basis. Concretely, given an inferred functional topology, we extracted the connected components and sought to match them with the known functional groups. The matching process was as follows (see Figure~\ref{fig:accuracy} for illustration). For each inferred connected component and for each known functional group we computed a per connected component and functional group F1 score (henceforth referred to as PCCFG-F1 score). The PCCFG-sensitivity was defined as the number of nodes of this particular functional group within this connected component divided by the size of the functional group; the PCCFG-precision was defined as the number of nodes of the functional group within this connected component divided by the size of the connected component. We identified each functional group to the connected component that maximised the resulting PCCFG-F1 score. As illustrated by Figure~\ref{fig:accuracy}, one connected component could be matched to more than one functional group.  Then, the overall precision for this matching was calculated as the ratio between the total number of nodes of a functional group represented in its matched connected component and the total size of the matched components; the overall sensitivity was calculated as the ratio between the total number of nodes of a functional group represented in its matched connected component and the total number of nodes in the functional connectivity (see Figure~\ref{fig:accuracy} for a worked out example). 

%%%%%%%%%%%%%%%%%%%%%%%%%%%%%%%%%%%%%%%%%%%%%%%%%%%%%%%%%%%%%%%%%%%%%%%%%%%%%%%%%%%%%%%%%%%%%%%%%%%%%%%%%%%%%%%%%%%%
\subsubsection{Description of benchmarking methods}\label{subsec:benchmark}\
%%%%%%%%%%%%%%%%%%%%%%%%%%%%%%%%%%%%%%%%%%%%%%%%%%%%%%%%%%%%%%%%%%%%%%%%%%%%%%%%%%%%%%%%%%%%%%%%%%%%%%%%%%%%%%%%%%%%

\noindent  Here, we briefly summarise the three methods used to benchmark our method. 

\noindent \textbf{Time-varying Graphical Lasso.} Hallac et al.~\cite{hallac2017network} extended the graphical Lasso algorithm and developed a method to solve for $\Theta = (\Theta_1,\Theta_2,...,\Theta_T)$ a set of symmetric positive definite matrices:
$$min_{\Theta \in S_{++}^{p}} \sum_{i=1}^{T} -l_i(\Theta_i) + \lambda ||\Theta_i|| + \beta \sum_{i=2}^{T} \Phi(\Theta_i - \Theta_{i-1}),$$
\noindent where  $T$ is the number of windows, $l_i(\Theta_i) = n_i(\text{log det} \Theta_i - Tr(S_i\Theta_i))$ is a function that encourages $\Theta_i$ to be close to $S_i^{-1}$ the inverse of the empirical covariance (if $S_i$ is invertible), $n_i$ is the number of observations, $||\Theta_i||$ is the semi-norm of $\Theta_i$, $\lambda$ is a positive constant that is adjusted to enforce the sparsity of the covariance matrix, $\Phi(\Theta_i - \Theta_{i-1})$ is a convex penalty function minimised at $\Phi(0)$, which encourages similarity between $\Theta_t$ and $\Theta_{t-1}$ and $\beta$ is a positive constant determining how strongly correlated neighbouring covariance estimations should be. The connectivity at time $t$ is then simply extracted from the non-zeros values of the precision matrix $\Theta_t$. Our results were obtained using their implementation available from \url{https://github.com/davidhallac/TVGL}.\newline

\noindent \textbf{Log Causality Inference.} The method by Kobayashi et al.~\cite{Kobayashi2017} assumes a direct acyclic graph (DAG) of events corresponding to the causality of events and proceeds in three steps. First they preprocess the data and remove events of time series that show strong temporal periodicity. Then, for every pair of nodes ($X$,$Y$), they state that it is not forming an edge if for at least one node $Z$, $X$ and $Y$ are conditionally independent ($P(X,Y|Z) \approx P(X|Z)P(Y|Z)$). Independence is tested using the conditional cross-entropy and the G-square test:
$$G^2 = 2mCE(X,Y|Z)$$
where $m$ is the duration of the recording. They furthermore define the direction of the edge using what they call the \textit{V-structure rule}. Finally, they post-process the data and remove frequently appearing edges to enable the detection of unusual important causality. Since they return topologies on a daily basis, this enables the identification of changes over time. Our results were obtained using their implementation available from \url{https://github.com/cpflat/LogCausalAnalysis}.\newline

\noindent \textbf{Correlation-based method.} We first bin all time series in bins of duration $\Delta_t$. They are thus of duration $D = \frac{T}{\Delta_t}$. For each pair of binned event time series ($x$,$y$) of length $D$, we computed the Pearson's correlation coefficient:
$$ r = \frac{\sum_{t=1}^{D} (x(t) - \mu_x)(y(t) - \mu_y)}{(D-1)\sigma_x\sigma_y}, $$
where $\mu_x$, $\mu_y$, $\sigma_x$ and $\sigma_y$ are the mean and standard deviation of the binned time series $x$ and $y$ respectively. To permit statistical testing, we followed the classical approach of applying the Fisher's $z$-transformation: 
$$ z = \frac{ln(1-r)}{ln(1+r)}. $$ Under the null hypothesis that the time series are independent, z should be asymptotically Gaussian with mean $0$ and standard deviation $\sigma_z = \frac{1}{\sqrt{D-3}}$. We then add an edge between $x$ and $y$ if $z > \alpha \times \sigma_z$, where $\alpha$ is a threshold parameter.

%%%%%%%%%%%%%%%%%%%%%%%%%%%%%%%%%%%%%%%%%%%%%%%%%%%%%%%%%%%%%%%%%%%%%%%%%%%%%%%%%%%%%%%%%%%%%%%%%%%%%%%%%%%%%%%%%%%%
\subsubsection{Results}\label{subsec:comparison}\
%%%%%%%%%%%%%%%%%%%%%%%%%%%%%%%%%%%%%%%%%%%%%%%%%%%%%%%%%%%%%%%%%%%%%%%%%%%%%%%%%%%%%%%%%%%%%%%%%%%%%%%%%%%%%%%%%%%%

\begin{figure}[h]
\begin{center}
	\includegraphics[width=0.78\linewidth]{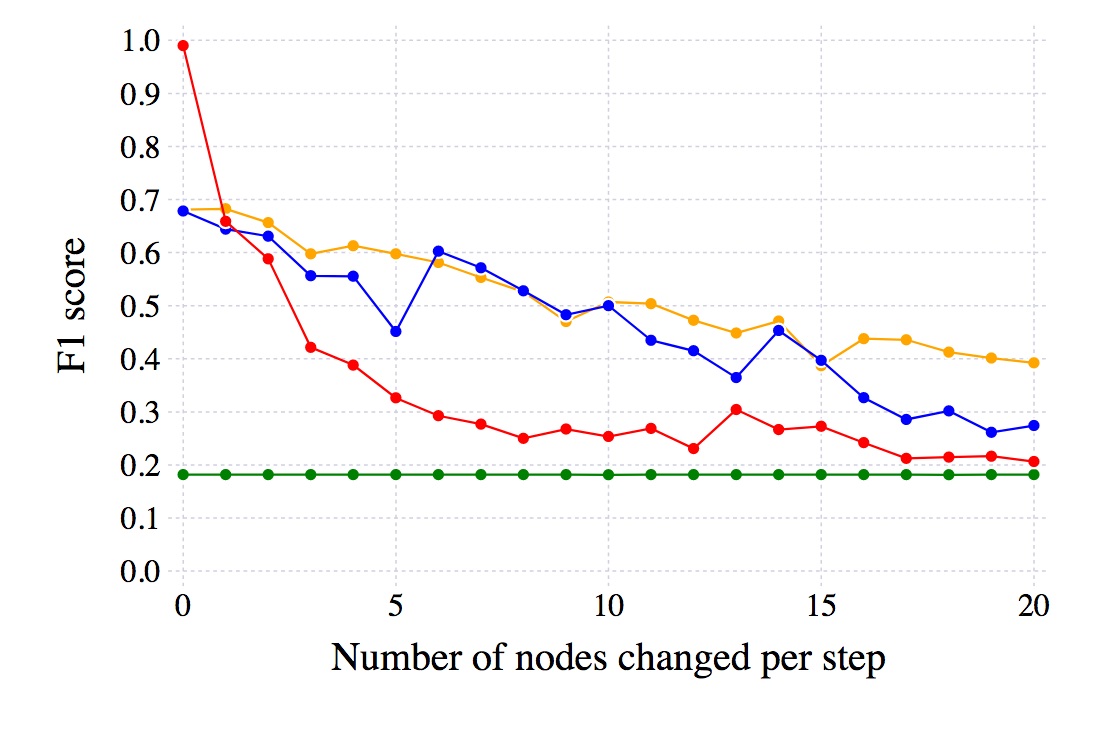}
	\caption{Evolution of the F1 scores for the four methods (blue: proposed method; orange: Kobayashi et al.; red: correlation-based method; green: Hallac et al.) for the static case (first data point = 0 nodes changed per step) and for when the functional connectivity is changed at increasingly higher rates.}
	\label{fig:res_change}
	\end{center}
\end{figure}

\noindent To systematically compare the performance of each method, we defined a reference set of parameter values which were then systematically varied under two conditions. The choice of these values was strongly affected by the lack of scalability of the benchmark methods and the need to keep calculations within reasonable timeframes given the number of scenarios considered. 

The reference network consisted of 100 nodes partitioned in 10 functional groups. Each functional group was affected by 200 cascades over 10 days (with a sampling time of 1 second) and each cascade involved half of the nodes in the functional group. Thus each node emitted 100 events on average. The maximum delay between the onset of a cascade and an event was fixed to 60 seconds. 

Using this reference network, we first evaluated the four methods when the number of nodes changed at each of the 50 steps varied from 0 node per step (i.e., the static case, for reference) to 20 nodes per step (i.e., a highly dynamic network). With 100 nodes in the network, a change of 1 node per step corresponds to a 1\% change in the functional connectivity. Since both the proposed methodology and the correlation-based method rely on a threshold parameter, we used the values of probability threshold that \textit{a posteriori} maximised the F1 score for both methods (one threshold value for each method).  \newline

As shown by Figure~\ref{fig:res_change}, the correlation-based method is unsurprisingly performing best in the static case. As the rate of change increases, however, performance of the correlation-based method rapidly drops whilst our proposed methodology shows comparable performance to that of Kobayashi et al. As the rate of changes increases, performance of all three methods degrade as expected. In all cases, the Hallac et al. method returned a F1 score that corresponded to either a fully connected graph or a fully disconnected graph ($\frac{2 \times 1 \times 0.1}{1 + 0.1} = 0.18$). It is clear that such poor performance is purely the result of it being asked to apply outside its operational range. Indeed, in order to be able to produce results with our data, binning was necessary. Each bin was set to represent one tenth of a day. We then learned the $\lambda$ and $\beta$ parameters on the reference network and obtained  $\lambda=0$ and $\beta=5$, thus, putting the emphasis on edge discovery. This, in turn, appears to have prevented it from efficiently recovering the functional topology. Accordingly, it will not be included in what follows. 

The above results considered rates of changes that were not necessarily plausible given our application scenario so in what follows we considered two scenarios: slowly changing network (1\% change per step -- second data point in Figure~\ref{fig:res_change}), rapidly changing network (10\% change per step -- 11th data point in Figure~\ref{fig:res_change}). In both cases, we systematically varied the reference network's parameters. 

\begin{figure*}[pt]
\begin{center}
	\includegraphics[width=0.40\linewidth]{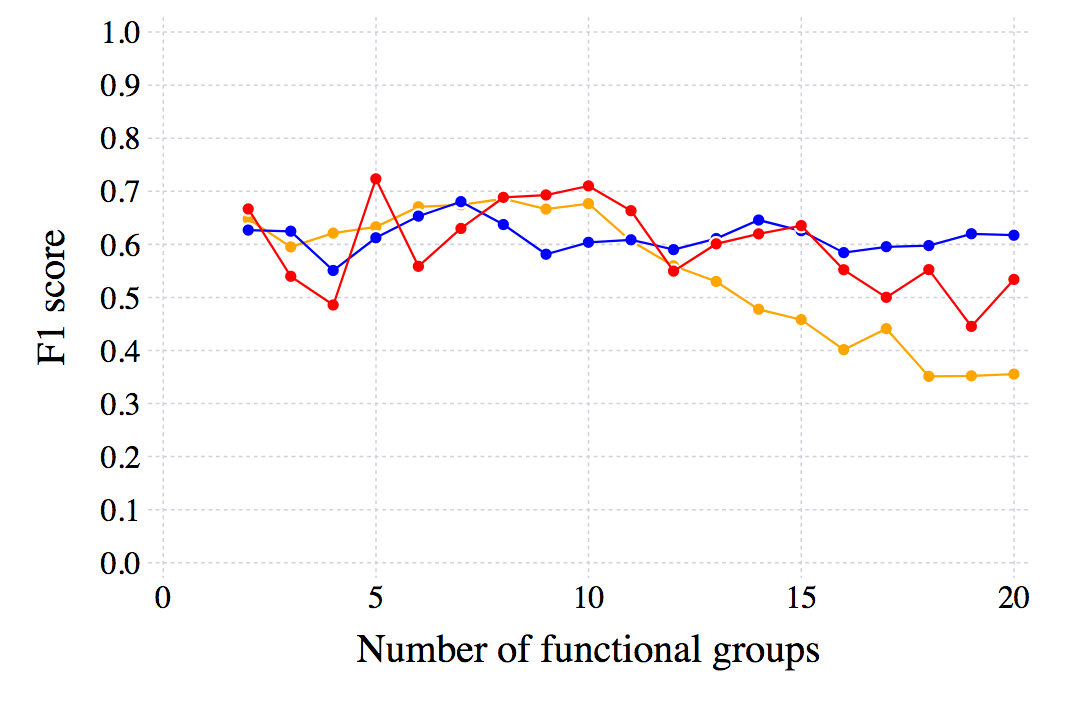}
	\hspace{0.5cm}
	\includegraphics[width=0.40\linewidth]{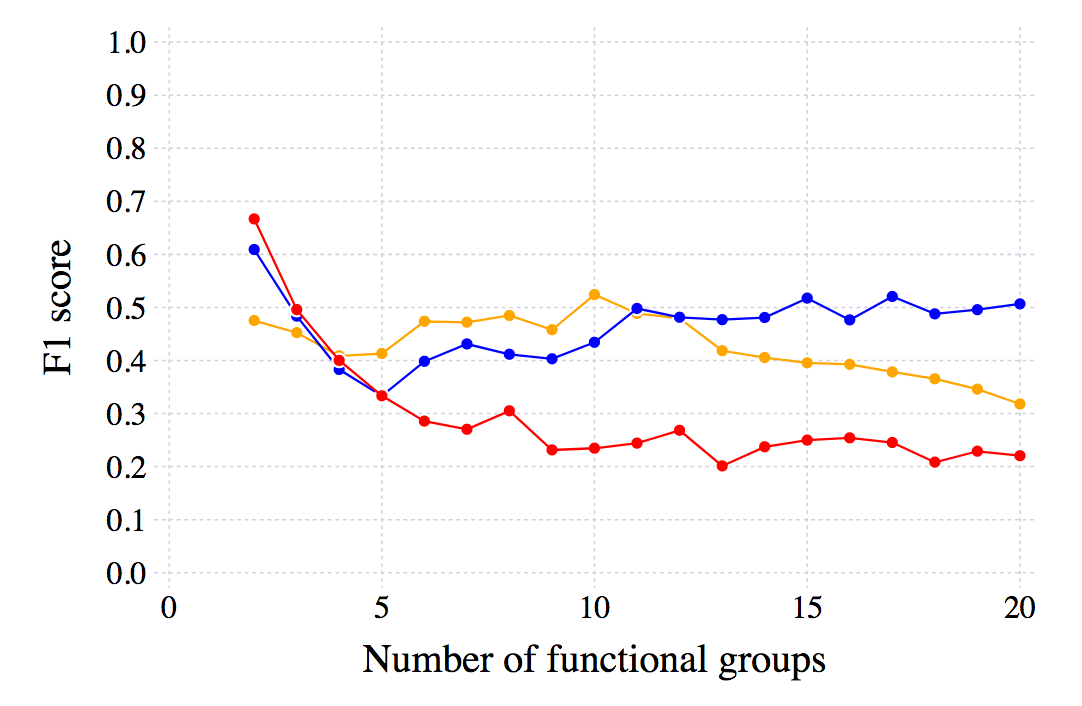}\\
	\includegraphics[width=0.40\linewidth]{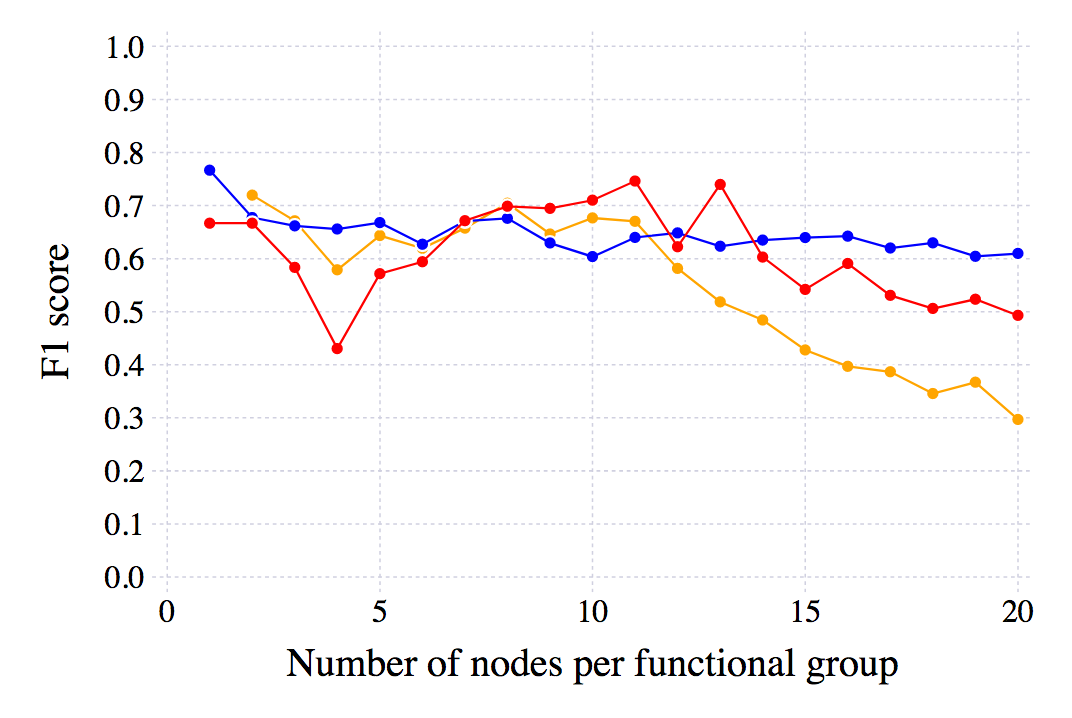}
	\hspace{0.5cm}
	\includegraphics[width=0.40\linewidth]{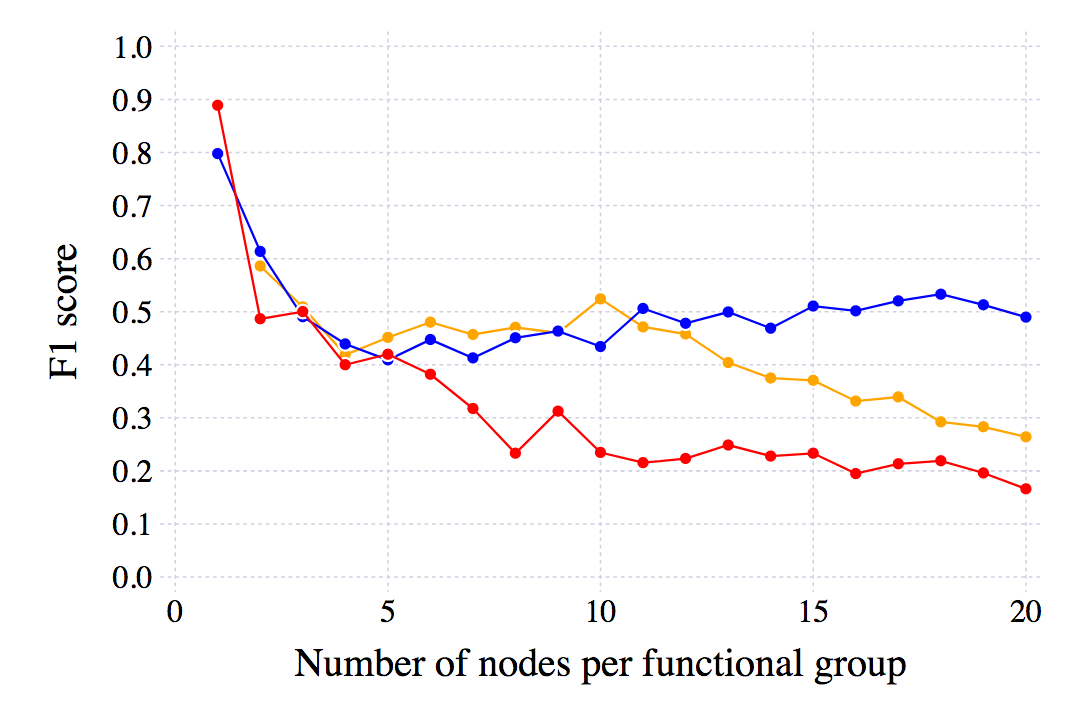}\\
    \caption{Evolution of the F1 scores for three of the four methods (blue: proposed method; orange: Kobayashi et al.; red: correlation-based method) when network size is systematically varied under two scenarios: slowly-changing network (left-hand side), rapidly-changing network (right-hand side). First row: Effect of an increase in the number of functional groups (each with 10 nodes). Second row: Effect of an increase in the number of nodes per functional group (there are 10 functional groups). See details in text.}
    \label{fig:res_synth1}
    \end{center}
\end{figure*}
    
\noindent  We first varied the size of the network via manipulation of the number of functional groups (each with 10 nodes) and that of the number of nodes per functional group (10 groups in all cases). As shown by Figure~\ref{fig:res_synth1}, as the network size increases, the performance of both the correlation-based method and that of Kobayashi et al. drops, even though the amount of information remains stable since the number of events emitted by each node is unchanged. In contrast, the proposed methodology shows a relatively stable F1 score. Crucially, even though all methods perform worse when the network changes at a faster rate, the above qualitative observation holds as far as the proposed methodology is concerned. The difference in accuracy between the correlation-based method and that of Kobayashi et al. stems from the fact that the former does not select enough edges while the latter selects too many. For two nodes $X$ and $Y$ that are part of the same functional group, when the size of the network increases, it becomes increasingly likely to find a node $Z$ that would be spuriously correlated to either or both of $X$ and $Y$. The method of Kobayashi et al. would see conditional independence and therefore remove an edge, whereas the correlation-based method would add an edge connecting those three nodes. \newline

\begin{figure*}[pt]	
	\begin{center}
	\includegraphics[width=0.40\linewidth]{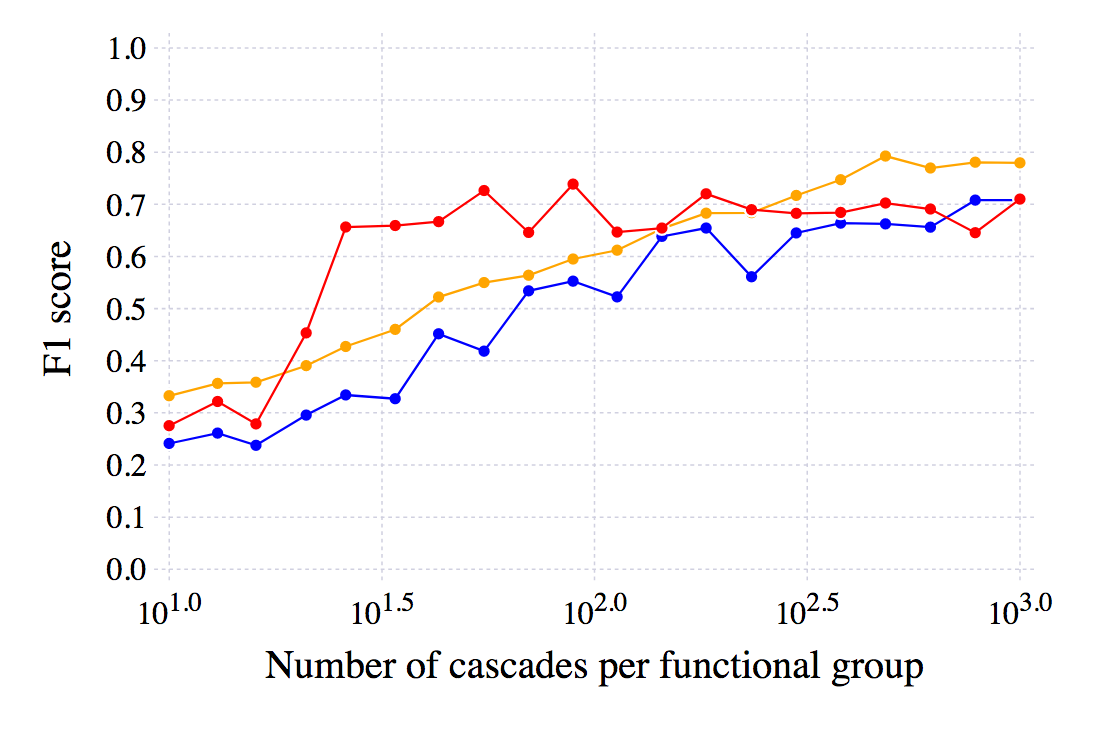}
	\hspace{0.5cm}
	\includegraphics[width=0.40\linewidth]{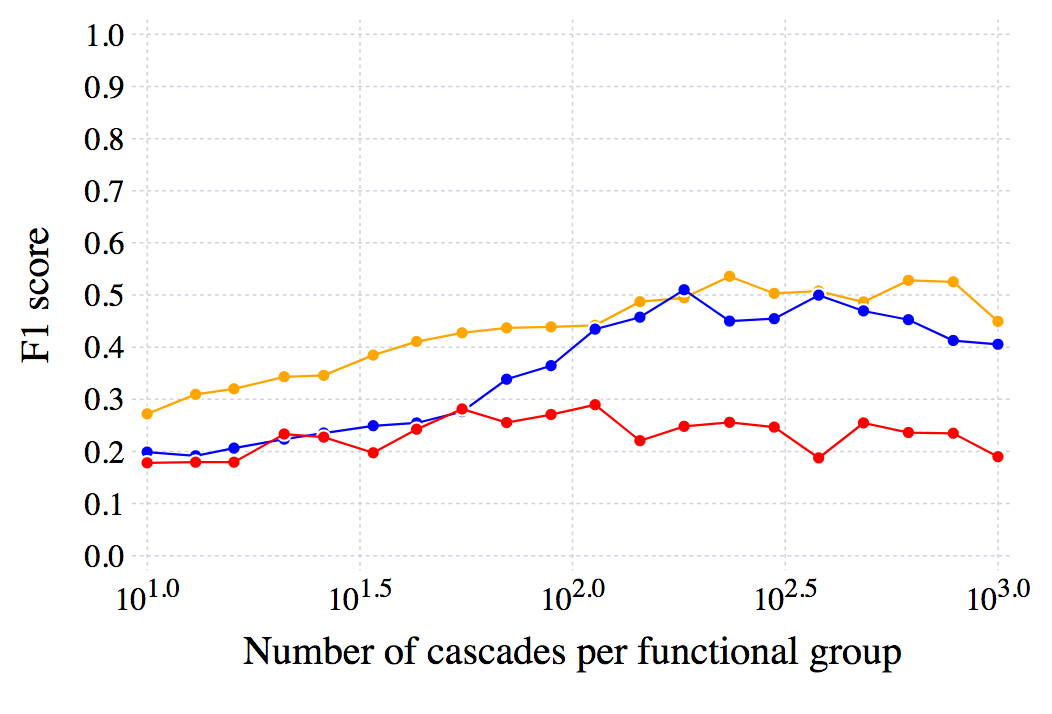}\\
	\includegraphics[width=0.40\linewidth]{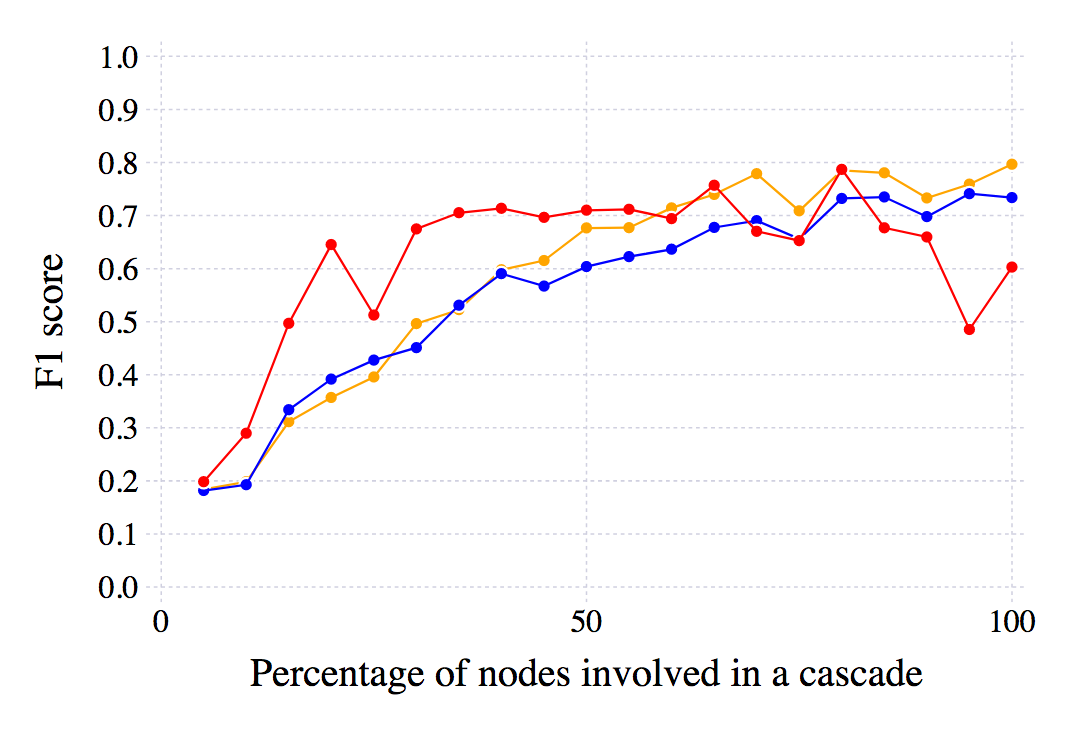}
	\hspace{0.5cm}
	\includegraphics[width=0.40\linewidth]{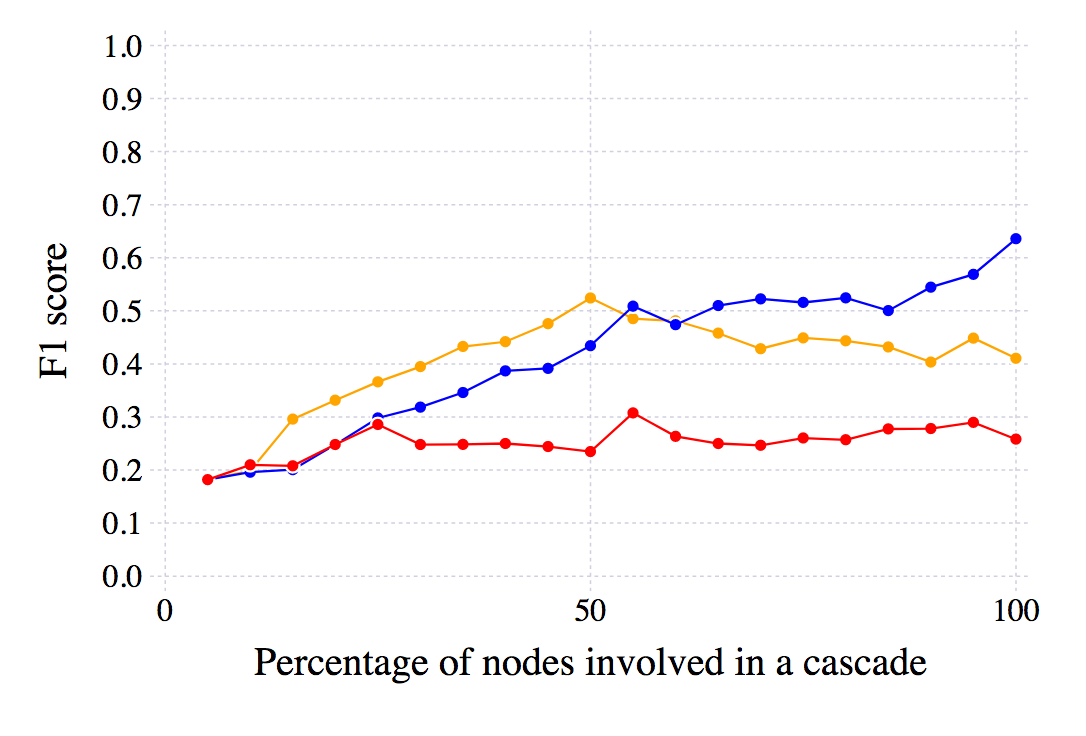}\\
	\includegraphics[width=0.40\linewidth]{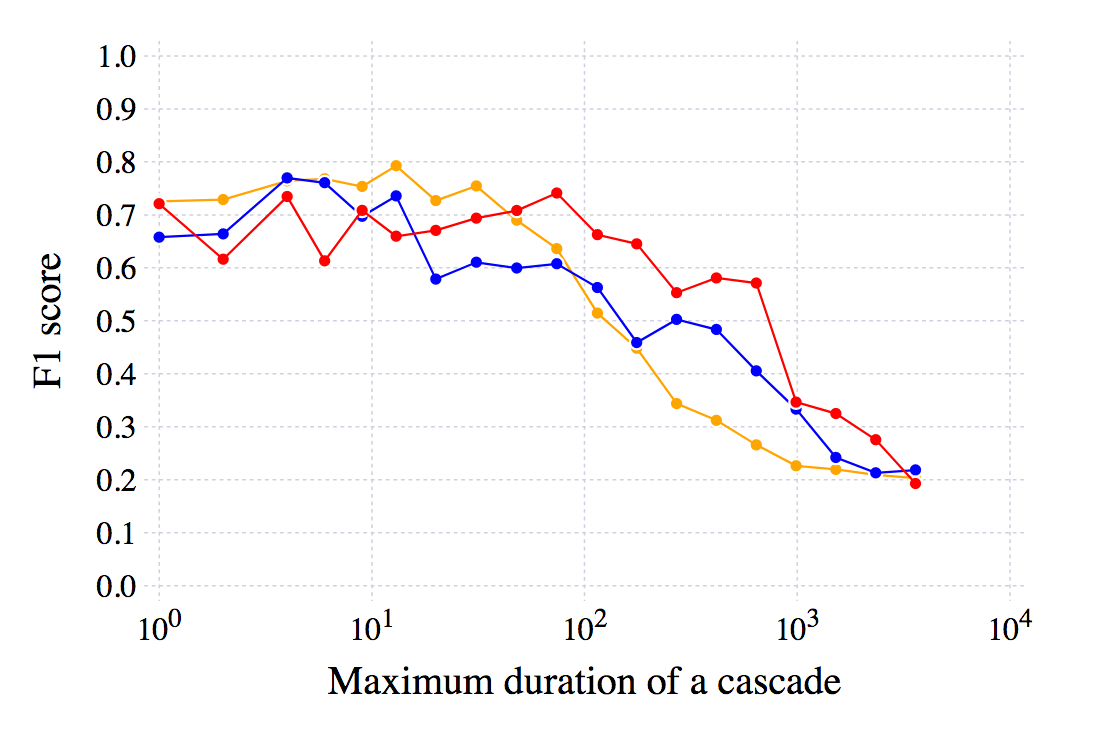}
	\hspace{0.5cm}
	\includegraphics[width=0.40\linewidth]{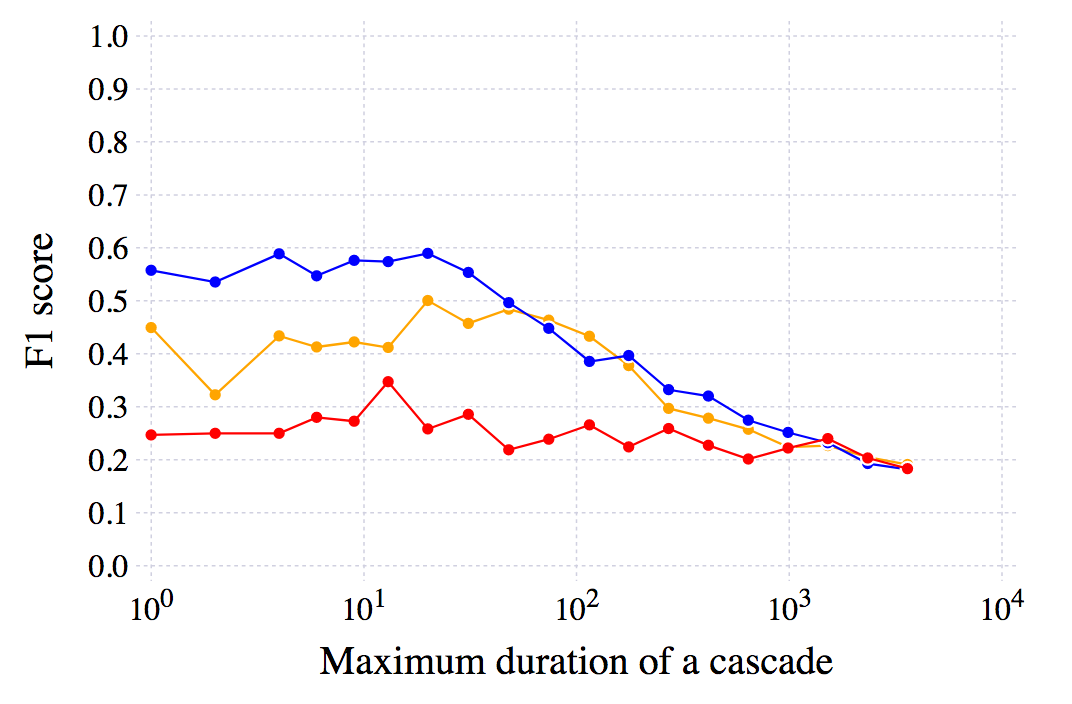}\\
    \caption{Evolution of the F1 scores for three of the four methods (blue: proposed method; orange: Kobayashi et al.; red: correlation-based method) when cascade-related parameters are systematically varied under two scenarios: slowly-changing network (left column), rapidly-changing network (right column). See details in text.}
   \label{fig:res_synth2}
   \end{center}
\end{figure*}
	
\noindent We next varied event frequency along two dimensions, either by increasing the number of cascades per functional group or by increasing the percentage of nodes involved in each cascade (first two rows in Figure~\ref{fig:res_synth2}). Here again, there is a general drop in performance across all methods as the rate of change in the network increases. However, there is a noticeable difference in regard to the ability of the correlation-based method to deal with a time-varying connectivity. An increase in the number of cascades corresponds to an increase in the density of events in the record. Whilst in the near-static case all three methods show similar behaviour, in the more dynamic case, the proposed methodology and that of Kobayashi et al. still show an increase (as expected) whereas the correlation-based method flatlines. The increase in the percentage of nodes involved in a cascade tests the sensitivity of the methods to how coherent the time series of the nodes are (where coherence reflects the likelihood of nodes emitting events in synchrony). In the near-static case, all three methods perform similarly. In the more dynamic case, however, once again the correlation-based method flatlines whereas the proposed method and that of Kobayashi et al. show a similar rate of increase, thus suggesting a similar information capture mechanism. \newline

\noindent Finally, we tested the sensitivity of the methods to the temporal extent of cascades, i.e., the maximum latency between the onset of a cascade and the latest event resulting from it. For a given rate of occurrence of cascade, an increased latency increases the likelihood of overlap between cascades and therefore not surprisingly performance drops significantly for all three methods in the near-static case. In the more dynamic case, the correlation-based method perform poorly irrespective of duration whereas the proposed methodology (and that of Kobayashi et al. to a lesser extent) shows some robustness over a wider range of values. 

%%%%%%%%%%%%%%%%%%%%%%%%%%%%%%%%%%%%%%%%%%%%%%%%%%%%%%%%%%%%%%%%%%%%%%%%%%%%%%%%%%%%%%%%%%%%%%%%%%%%%%%%%%%%%%%%%%%%
\subsection{Application to real-world computer network data}\label{sec::RWapp}
%%%%%%%%%%%%%%%%%%%%%%%%%%%%%%%%%%%%%%%%%%%%%%%%%%%%%%%%%%%%%%%%%%%%%%%%%%%%%%%%%%%%%%%%%%%%%%%%%%%%%%%%%%%%%%%%%%%%

%%%%%%%%%%%%%%%%%%%%%%%%%%%%%%%%%%%%%%%%%%%%%%%%%%%%%%%%%%%%%%%%%%%%%%%%%%%%%%%%%%%%%%%%%%%%%%%%%%%%%%%%%%%%%%%%%%%%
\subsubsection{Data description}\label{subsec:rwds}\
%%%%%%%%%%%%%%%%%%%%%%%%%%%%%%%%%%%%%%%%%%%%%%%%%%%%%%%%%%%%%%%%%%%%%%%%%%%%%%%%%%%%%%%%%%%%%%%%%%%%%%%%%%%%%%%%%%%%

\noindent The authors were provided with 5-month worth of event logs of a large Internet Service Provider. From these logs, we extracted time-series of events emitted by devices on their network (recorded with a sampling period of 1 second). In total, there were 473,580 events distributed over 53,604 devices. Every event is accompanied by a type field which usually contains a single word. This field is mapped to specific types of events at deployment time. Such mapping can be manual, e.g. for events received from third-party event management/monitoring systems, or automatic, i.e., involving text processing to extract meaningful types from raw log messages. Out of a total of 41 distinct types, 8 are found to account for over 90\% of all events in the record. Analysis of the data (in submission) suggested that some nodes tend to emit events of the same type more often than not, thus suggesting they form part of a functional topology. This will be the basis for our analysis below. 

%%%%%%%%%%%%%%%%%%%%%%%%%%%%%%%%%%%%%%%%%%%%%%%%%%%%%%%%%%%%%%%%%%%%%%%%%%%%%%%%%%%%%%%%%%%%%%%%%%%%%%%%%%%%%%%%%%%%
\subsubsection{Results}
%%%%%%%%%%%%%%%%%%%%%%%%%%%%%%%%%%%%%%%%%%%%%%%%%%%%%%%%%%%%%%%%%%%%%%%%%%%%%%%%%%%%%%%%%%%%%%%%%%%%%%%%%%%%%%%%%%%%

\begin{figure}[h]
\begin{center}
    \includegraphics[width=\linewidth]{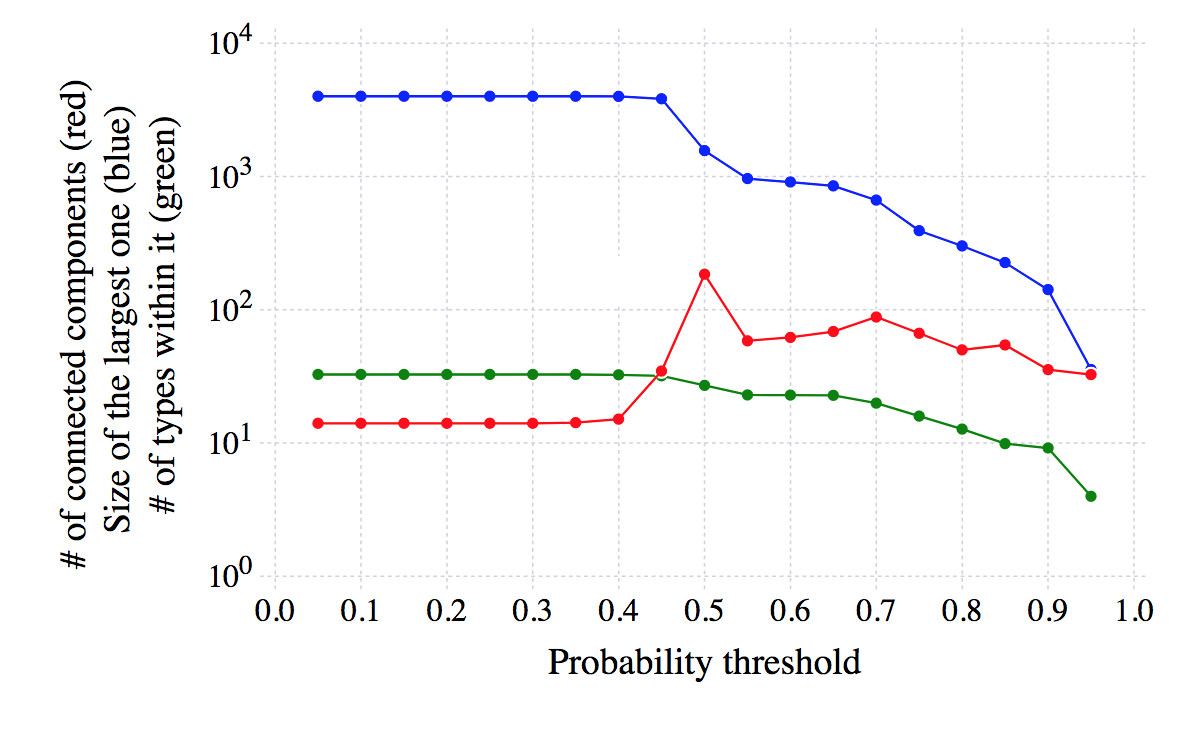}
    \caption{Effect of the probability threshold on the number connected components in the inferred topology, the size of the largest component and the number of event types represented within it.}
   \label{fig:typeingc}
   \end{center}
\end{figure}

\noindent We deployed the proposed methodology on the full network %(made of 53,604 devices emitting 473,580 alerts 
and analysed its inferred topology when systematically varying the probability threshold (used to determine the presence of a functional link) between 0.1 (very loose) and 0.9 (very strict). In general (see Figure~\ref{fig:typeingc}), our method returns a large connected component (e.g., 2,673 nodes and 33,928 edges for a probability threshold of 0.5) as well as a multitude of smaller connected components (e.g., 242 smaller connected components involving 942 nodes altogether for a probability threshold of 0.5).   
As the probability threshold increases, the percentage of devices (in those small components) that emit events of the same type increases (from 50\% for low thresholds to 80\% for the highest threshold). 

\noindent The largest connected component typically involves devices that emit events of various types. Nevertheless, as shown by Figure~\ref{fig:typeingc}, the number of types involved drops significantly with increasing probability threshold (note log scale). Further, an analysis of the average shortest path between devices emitting events of a same type tends to be smaller for some types than the average shortest path between random nodes. Taken together, our findings provide reasonable evidence of the existence of communities (i.e., different services) within the inferred topology. 

\section{Conclusion}

In this paper, we have described a new method to infer the functional connectivity of a large-scale computer network from sparse time series of its node events. We did so under three strong constraints: (a) non-stationarity of the functional connectivity due to unknown temporal changes in the network, (b) sparsity of the time-series of events that limited the effectiveness of classical correlation-based analysis, and (c) lack of an explicit model describing how events propagate through the network. This is a hard problem. When using synthetic data for which ground truth was available, the F1-score only rarely exceeded 0.7 in the near-static case, 0.6 in the more dynamic case. However, this should not detract from the fact that the method was able to recover a substantial amount of the connectivity, including its changes over time, from an extremely limited amount of information. Indeed, it did so at least as well as state of the art methods in the near-static case, and usually better in the dynamic case. Importantly, unlike existing network inference methods (that typically do not handle sparse data well), it remains computationally tractable even with large networks (here, 10,000 nodes) over very long records (here, $10^7$ observations). To be able to benchmark our method against state-of-the-art methods, we were reduced to networks of size magnitudes smaller than our real-world application. The critical lack of scalability of these methods cannot be overstated. For example, to be able to use the method by Hallac et al.~\cite{hallac2017network}, binning of the time series was required, which in turn, rendered the method non-operational (despite the small size of the networks). And, whereas our method could produce results for the various scenarios involving the synthetic data in $20$ minutes on a single laptop, the method by Kobayashi et al.~\cite{Kobayashi2017} required $30$ minutes on the HPC cluster with 72 nodes working in parallel. 

In principle, this method could be applied to any system in which functional relationships between nodes translate into short delays between their respective activities. To fulfil its applicative potential, however, a more complete understanding of the various assumptions and parameters underpinning the method must be obtained. In particular, a more rigorous understanding of the method's sensitivity to the choice of key parameters, e.g., number of windows and probability threshold to name just two, is warranted. Preliminary investigations suggest both impact the balance between precision and sensitivity and that in itself may prove to have value when different application scenarios are considered. On a related note, although the method was motivated by a particular application domain, we sought to remain as detached as possible from its specifics. However, it is unclear to what extent we were successful in doing so. To improve the scope of possible applications, a first interesting avenue for further development is to allow the model parameters to change over time. This would make it possible for the method to learn non constant rates of change of the connectivity for example and adapt to different connectivity changes. Another direction is to explore the extent to which relaxing some of the constraints may improve predictive power, e.g., using prior knowledge about how events propagate in the network.

\bibliography{biblio}
\bibliographystyle{plain}

\end{document}